\documentclass[letterpaper, 10 pt, journal, twoside]{IEEEtran}  %

\IEEEoverridecommandlockouts                              

\usepackage{graphicx} 
\usepackage{amsmath} 
\usepackage{amssymb}  
\usepackage[usenames, dvipsnames]{color}

\usepackage{lipsum}
\usepackage{color}
\usepackage{cite}

\usepackage{diagbox}
\usepackage{tabu}
\usepackage{balance}  

\usepackage[ruled,linesnumbered]{algorithm2e}
\usepackage{algpseudocode}
\usepackage{epsfig}

\newcommand{\revision}[1]{\textcolor{black}{#1}} 
\newcommand{\secrevision}[1]{\textcolor{black}{#1}} 

\title{
Precise Repositioning of Robotic Ultrasound: Improving Registration-based Motion Compensation using Ultrasound Confidence Optimization
}

\author{Zhongliang Jiang$^{1}$, \textit{Member, IEEE}, Nehil Danis$^{1}$, Yuan Bi$^{1}$, Mingchuan Zhou$^{2}$,\\ Markus Kroenke$^{3}$, Thomas Wendler$^{1}$, and Nassir Navab$^{1,4}$, \textit{Fellow, IEEE} 
\thanks{This study was supported in part by ZJU 100 Young Talent Program. Zhongliang Jiang and Nehil Danis contributed equally to this work.}
\thanks{$^{1}$Z. Jiang, N. Danis, Y. Bi, T. Wendler, and N. Navab are with the Chair for Computer Aided Medical Procedures and Augmented Reality, Technical University of Munich, Garching, Germany. {\tt\footnotesize{(zl.jiang@tum.de)}}
        }%
\thanks{$^{2}$M. Zhou (\textit{Corresponding author}) is with the College of Biosystems Engineering and Food Science, Zhejiang University, Hangzhou, China.}
\thanks{$^{3}$M. Kroenke is with the Department of Radiology and Nuclear Medicine, German Heart Center Munich, Technical University of Munich, Germany.}
\thanks{$^{4}$N. Navab is also with the Laboratory for Computational Sensing and Robotics, Johns Hopkins University, Baltimore, MD, USA.}
}

\begin{document}

\maketitle


\begin{abstract}
Robotic ultrasound (US) imaging has been seen as a promising solution to overcome the limitations of free-hand US examinations, i.e., inter-operator variability. \revision{However, the fact that robotic US systems cannot react to subject movements during scans limits their clinical acceptance.} Regarding human sonographers, they often react to patient movements by repositioning the probe or even restarting the acquisition, in particular for the scans of anatomies with long structures like limb arteries. To realize this characteristic, we proposed a vision-based system to monitor the subject's movement and automatically update the scan trajectory thus seamlessly obtaining a complete 3D image of the target anatomy. The motion monitoring module is developed using the segmented object masks from RGB images. Once the subject is moved, the robot will stop and recompute a suitable trajectory by registering the surface point clouds of the object obtained before and after the movement using the iterative closest point algorithm. Afterward, to ensure optimal contact conditions after repositioning US probe, a confidence-based fine-tuning process is used to avoid potential gaps between the probe and contact surface. Finally, the whole system is validated on a human-like arm phantom with an uneven surface, while the object segmentation network is also validated on volunteers. The results demonstrate that the presented system can react to object movements and reliably provide accurate 3D images. 
\end{abstract}


\markboth{IEEE Transactions on Instrumentation and Measurement. Accepted August, 2022}
{Jiang \MakeLowercase{\textit{et al.}}: Precise Repositioning of Robotic Ultrasound}




\bstctlcite{IEEEexample:BSTcontrol}
\section{Introduction}
\IEEEPARstart{U}{ltrasound} (US) imaging is a powerful and effective tool for the diagnosis of lesions and abnormalities of internal tissues and organs. US imaging is non-invasive, real-time, radiation-free and widely available. In 2017, over $9.2$ million US scans were performed in England. This number is twice and three times larger than the number of computer tomography (CT) scans and magnetic resonance imaging (MRI), respectively, during the same period~\cite{hoskins2019diagnostic}. Besides, US has been employed as the primary mechanism to diagnose peripheral artery disease (PAD) in clinical practice~\cite{aboyans20182017}. Regarding the femoral artery, Collins~\emph{et al.} reported that US imaging has $80\%$-$98\%$ sensitivity in detecting arterial stenoses~\cite{collins2007systematic}.

\begin{figure}[ht!]
\centering
\includegraphics[width=0.45\textwidth]{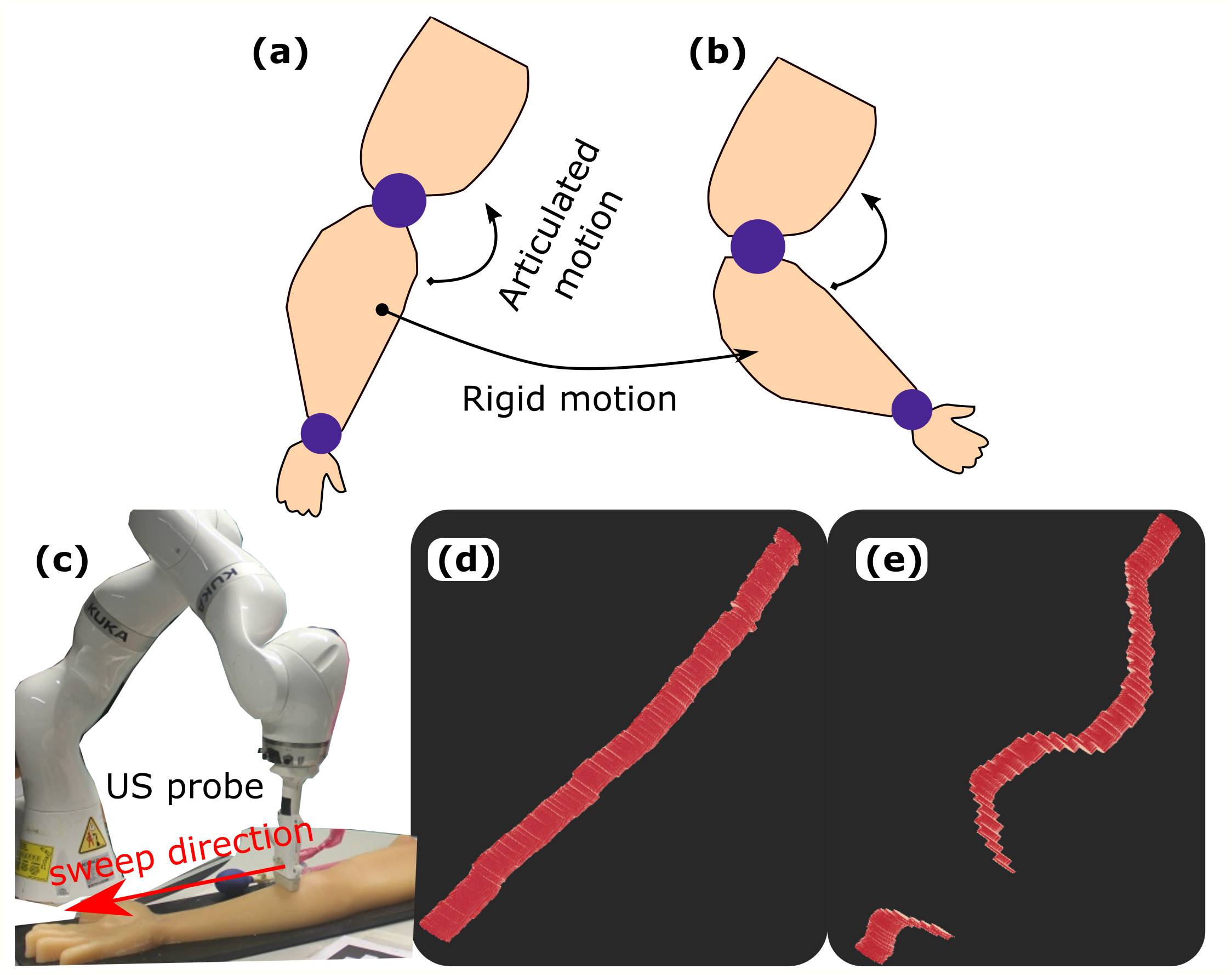}
\caption{Illustration of two types of motion of limbs and the impact of object movement on 3D US compounding. 
(a) and (b) are two poses of an arm. There is articulated motion between joints and rigid motion of the separated forearm. (c) US scan of a human-like arm phantom with an uneven surface. (d) 3D reconstruction result of an underlying vessel when the imaged object is stationary during the sweep, and (e) 3D result of the exact same vessel obtained when the object is moved randomly relative to the initial trajectory during the sweep. 
}
\label{Fig_shift}
\end{figure}

\par
However, due to the inherited limitations of the US modality, the conventional two-dimensional (2D) B-mode images suffer from high intra- and inter-operator variability. These variations result in inconsistent diagnosis results of examinations carried by sonographers with different level of experience, or even by the same operator at different times. To alleviate this problem, three-dimensional (3D) imaging is employed to intuitively characterize and quantify the abnormality, i.e., vascular stenosis~\cite{fenster20063d, merouche2015robotic}. Compared with 2D images, 3D images are more reproducible because they do not require exactly repeating the probe orientation. Ottacher~\emph{et al.} first introduced 3D US image to accurately guide the process for insertion of pedicle screws~\cite{ottacher2020positional}.

\par
To augment 2D images into 3D volumes, Hossack~\emph{et al.} designed a probe with a 2D arrangement of transducer elements instead of traditional 1D arrays~\cite{hossack2002quantitative}. The 2D array allows to directly visualize the objects of interest in 3D, while the cost of such systems is high. In addition, the size of the acoustic window is relatively small. In contrast, an optical or electromagnetic tracking system is used as an alternative way to obtain 3D volumes by providing tracked 2D images~\cite{gee2003engineering}. However, potential magnetic interference and occlusion between the optical camera and the markers limit the usability of such systems in real scenarios. To address this, Yang~\emph{et al.} employed a robotic manipulator to automatically produce 3D images of human spine~\cite{yang2021automatic}.  

\par
To further address this challenge, robotic techniques are employed to develop robotic US systems (RUSS) for accurate and reproducible US acquisitions. 
Gilbertson~\emph{et al.} developed an impedance controller for an one degree of freedom (DoF) device to stabilize US imaging during scans~\cite{gilbertson2015force}. Huang~\emph{et al.} attached two thin strain sensors on the front face of the US probe to maintain a constant force during scans~\cite{huang2018fully, huang2018robotic}. Besides contact force, Jiang~\emph{et al.} quantitatively optimize the orientation using both US imaging and estimated force at the end-effector to improve the imaging quality~\cite{jiang2020automatic}. Yet, this approach is only valid for a convex probe. To further accurately orient both linear and convex probes to unknown constraint surfaces, Jiang~\emph{et al.} further developed a mechanical model-based approach using the accurate force measurements from an external force sensor~\cite{jiang2020automaticTIE}. 
In addition, Huang~\emph{et al.} used a depth camera to compute the normal direction of the target surface for autonomous US scans~\cite{huang2018robotic}. Although the accuracy of the camera-based approach is not as high as the force-based approach, the estimated orientation can be computed in real-time. Additionally, benefited from force measurements, Sun~\emph{et al.}~\cite{sun2010trajectory} and Jiang~\emph{et al.}~\cite{jiang2021deformation} proposed approaches to further recover the force-induced deformation in 2D and 3D images, respectively.

\par
To automatically perform the screening of aorta by RUSS, Virga~\emph{et al.} computed a scan trajectory from a preoperative MR image and further transferred it to the robotic frame by registering the MR to the object surface captured by an RGB-D camera~\cite{virga2016automatic}. \revision{Nevertheless, potential movements of objects during scans haven't been considered, which may lead to failure of visualizing target anatomies. Considering the task to display long structures, e.g., a limb artery tree, sonographers even need to actively adjust the limb pose to overcome the limitation of robotic working space and to better visualize the target artery. Since the limbs are flexible, articulated motion will happen around the joints (see Fig.~\ref{Fig_shift}). However, the motion of separate parts, i.e., forearm, and upper arm, can be seen as rigid transformation. If the potential motion (articulated or rigid motion) is not considered during scans, the imaging quality will be significantly decayed in presence of motion as Fig.~\ref{Fig_shift} (e).} To address this, Jiang~\emph{et al.} developed a marker-based method to monitor and compensate for potential motion during scanning~\cite{jiang2021motion}. \revision{Yet, this approach relied on optical markers, which limits the usability of the approach in real scenarios. In addition, it is time-consuming to carefully configure the markers on individual patients.} 

\par
\revision{To further tackle this challenging}, we proposed an \revision{surface registration-based} motion-aware RUSS with the ability to reposition the probe for precise and complete 3D images of target anatomies in presence of rigid motions during scanning. This system combines the advantages of free-hand US (flexibility) and RUSS (accuracy and stability). To enable the motion-aware ability, an RGB-D camera is employed to monitor and compensate for potential object movements. 
This technique is particularly important for the development of robotic screening of PAD on limb arteries, where sonographers usually need to actively adjust the limb position to avoid exceeding working space limits and to visualize the whole artery tree. 
The main contributions are summarized as follows:

\begin{itemize}
  
  \item \revision{A learning-based approach is developed to monitor the object’s movement and automatically update the initial trajectory based on the surface registration technique to accomplish the scan for a complete and accurate 3D geometry. Compared with the state-of-the-art work~\cite{jiang2021motion}, the proposed surface registration-based approach is more compatible with the current clinical practices without the requirements of time-consuming tuning of patient-specific configurations.
  }
  
  
  \item \revision{An online probe orientation adjustment method is developed based on the US confidence map~\cite{karamalis2012ultrasound} to optimize the contact condition between probe and object surface during scans. Besides the aim of guaranteeing the image quality, this adjustment also benefits the visibility of the target object in the imaging view against surface registration errors. }
\end{itemize}
The segmentation approach are validated on both human arms and an arm phantom. In addition, the proposed motion-aware RUSS is fully validated on the human-like arm phantom with an uneven surface. The video\footnote{Video: https://www.youtube.com/watch?v=MUtgSXS7EZI} can be publicly accessed.




\section{System Overview}
\par
In this section, we first describe the overview of the proposed motion-aware RUSS for achieving an accurate and complete 3D volume of target anatomies against potential (expected or unexpected) object movements during the scanning (Section II-A). Then, the system calibration procedures and the compliant controller used to maintain a given force throughout US sweep are described in Section II-B and II-C, respectively.


\begin{figure*}[ht!]
\centering
\includegraphics[width=0.90\textwidth]{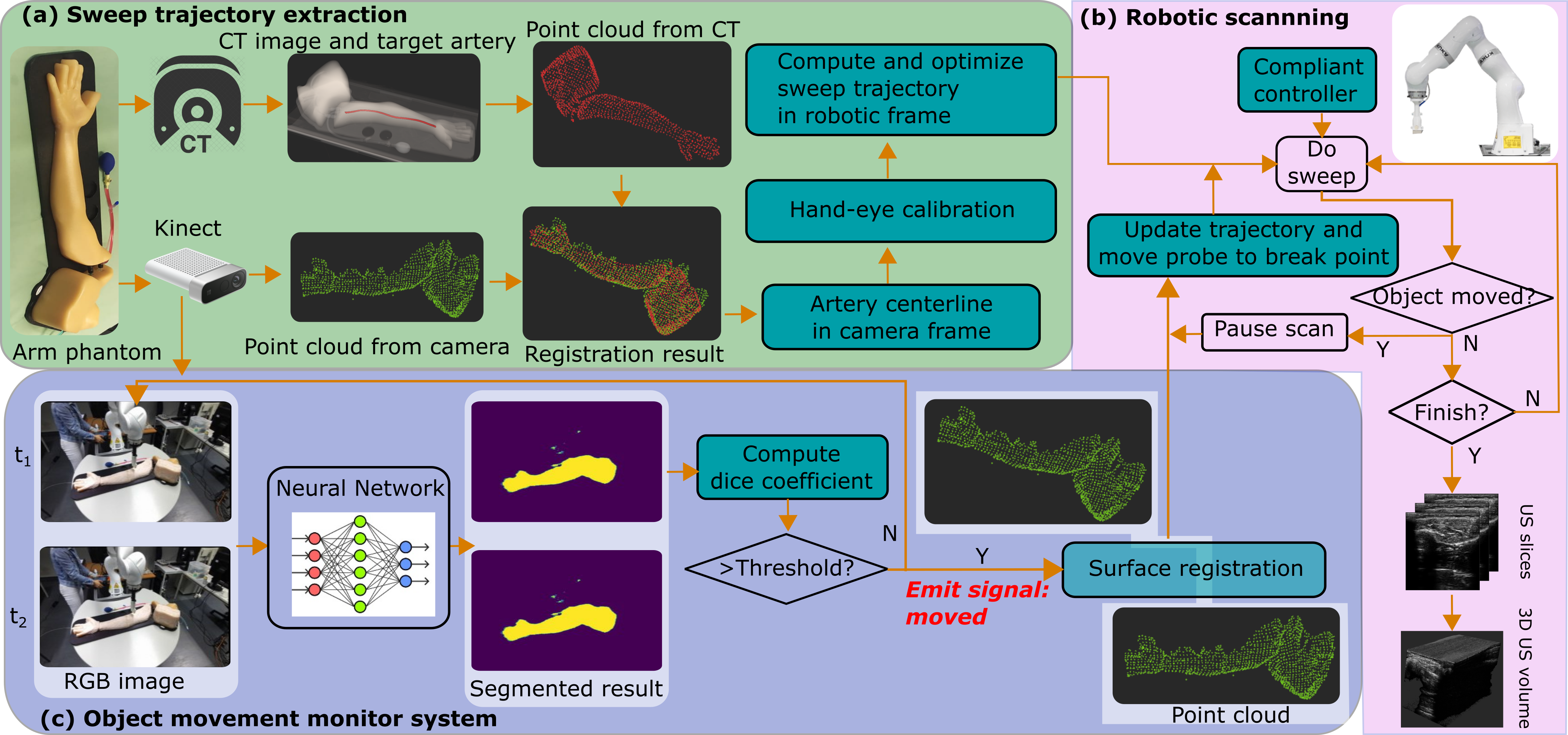}
\caption{\revision{The pipeline of the motion-aware RUSS.} (a) the sweep trajectory extraction and optimization module, (b) the robotic scanning execution module, and (c) the camera-based movement monitoring module. The dice threshold is $0.95$ in this work.}
\label{Fig_workflow}
\end{figure*}

\subsection{Workflow}
\par
To enable the motion-aware ability, a vision-based approach is developed in this study to detect the (expected or unexpected) movements and update the sweep trajectory based on the surface registration approach. The general pipeline is described in Fig.~\ref{Fig_workflow}. Compared with the marker-based approach~\cite{jiang2021motion}, the passive markers are not needed anymore and a human-like arm phantom is used to mimic real scenarios.


\subsubsection{Hardware Setup}
\par
The proposed motion-aware RUSS mainly consists of three components: a redundant robotic arm (LBR iiwa 14 R820, KUKA GmbH, Germany), a US machine (Cephasonics, USA), and an RGB-D camera (Azure Kinect, Microsoft Corporation, USA). \revision{The robotic arm has seven joints, and external joint torque sensors are integrated into all joints. This configuration enables the accurate control of contact force at the end-effector.} \revision{To have an overview of the whole setup, the camera is fixed in front of the robotic and objects of interest (arm phantom and volunteer's arm).} \revision{The resolutions of RGB camera and depth camera are set to $2048 \times 1536$ and $1024\times 1024$, respectively.}

\par
The robot is controlled using a self-developed Robot Operating System (ROS) interface~\cite{hennersperger2016towards}. The control commands are updated at $100~Hz$ to guarantee real-time performance. \revision{A linear probe (CPLA12875, Cephasonics) is attached to the robotic flange using a custom-designed holder. The B-mode images are accessed via a USB interface ($50~fps$) and visualized in a software platform (ImFusion Suite, ImFusion GmbH, Germany). The detailed configurations of US acquisition are listed as follows: imaging depth:~$55~mm$, brightness:~$67~dB$, frequency:~$7.6~MHz$.} To validate the proposed system, a commercial human-like arm phantom (PICC US Training Model, Skills Med Deutschland GmbH, Germany) is employed.

\subsubsection{Implementation Pipeline}
\par
\revision{
The motion-aware RUSS is realized by three main components: the sweep trajectory extraction and optimization module, the robotic scanning execution module, and the camera-based movement monitoring module. Regarding the extraction of sweep trajectory, the artery of interest is manually segmented from preoperative CT data using an open-source software 3D slicer\footnote{https://www.slicer.org/}. Then, the meshes of the arm surface and the target limb artery surface are generated and further used to create point clouds using Meshlab\footnote{https://www.meshlab.net/} [see Fig.~\ref{Fig_workflow}~(a)]. To transfer the target artery position from CT data to the robotic base frame, a self-occlusion point cloud of the arm surface is generated based on the RGB-D camera placed in front of the target object. After the registration between the two surface point clouds, the artery trajectory can be transferred into the camera frame. Further utilizing the hand-eye calibration result, the artery trajectory can be transferred into the robotic base frame to guide the probe movement for autonomous scans.} 

\par
\revision{
Afterward, the RUSS starts executing US examination along the artery. To provide good imaging quality and guarantee patient safety during the scanning, a compliant controller is employed to maintain a constant force between the probe and the contact surface~\cite{jiang2021autonomous, hennersperger2016towards}. In addition, to monitor the object movement, a neural network (UNet-VGG16~\cite{balakrishna2018automatic}) was trained to segment the arm from RGB images. The segmented results of images acquired at $t_1$ and $t_2$ are further used to compute the dice coefficient. Once the dice coefficient is smaller than a preset threshold, the system considers the object has moved, and a corresponding signal is emitted to the ROS master. After detecting such a signal, the robot stops, and the registration between the camera point clouds acquired at $t_1$ and $t_2$ is carried out. Based on the registration result, the remaining trajectory can be updated, and RUSS can continue the scan by automatically moving the probe to the breaking point, namely, the probe position when the movement happens. Benefited from this compensation, a complete and accurate 3D compounding can still be achieved when the object moves during the scans.}

\subsection{Coordinate System Transformation}
\par
\begin{figure}[ht!]
\centering
\includegraphics[width=0.48\textwidth]{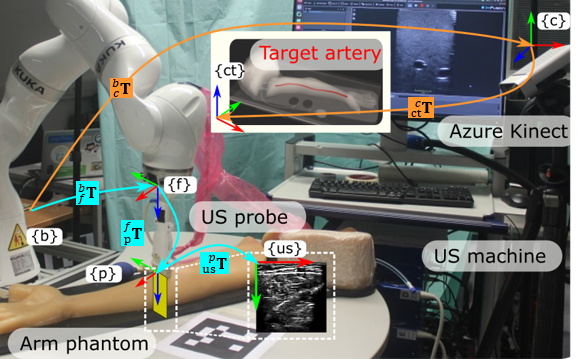}
\caption{Illustration of involved coordinate frames. Red arrows, green arrows and blue arrows represent X, Y and Z directions of corresponding coordinate systems, respectively. 
}
\label{Fig_coordinate_system}
\end{figure}

There are two different types of calibration procedures involved in this study: 1) US calibration and 2) hand-eye calibration. The former one is used to transfer the pixel position in B-mode images to the robotic base frame while the latter one is used to transfer the position from the camera view to the robotic base frame to guide robotic movement. The involved coordinate systems are depicted in Fig.~\ref{Fig_coordinate_system}: 1) the robotic base frame $\{b\}$, 2) the robotic flange frame $\{f\}$, 3) the probe frame $\{p\}$, 4) US imaging frame $\{us\}$, 5) the camera frame $\{c\}$, and 6) the CT atlas frame $\{ct\}$. 

\par
To perform trajectory correction based on US imaging feedback, the transformation matrix \revision{$^{b}_{us}\textbf{T}\in R^{4\times4}$} between frame $\{b\}$ and frame $\{us\}$ can be calculated as follows:

\begin{equation}\label{eq_trasformation}
^{b}_{us}\textbf{T} = ^{b}_{f}\textbf{T}~^{f}_{p}\textbf{T}~^{p}_{us}\textbf{T}
\end{equation}
where \revision{$^{j}_{i}\textbf{T}\in R^{4\times4}$} is the transformation matrix used to transfer the position from frame $\{i\}$ to frame $\{j\}$. $^{b}_{f}\textbf{T}$ represents the kinematic model of the robotic arm. This can be directly accessed by the API function provided by the manufacturer. Besides, $^{f}_{p}\textbf{T}$ depends on custom configuration. In this work, the rotation part of $^{f}_{p}\textbf{T}$ is set to an identity matrix $\textbf{I}^{3\times 3}$ and the translational part is obtained from the custom-designed 3D holder model. Due to the characteristic of linear probe that US elements are physically distributed on the probe tip within a specific length $L_p$, the horizontal mapping ($\frac{L_p}{W_{us}}$) between pixel-wise positions and physical positions can be computed, where $W_{us}$ is the pixel-wise width of B-mode imaging. Similarly, the vertical mapping is computed by $\frac{D}{H_{us}}$, where $D$ and $H_{us}$ are the depth setting and the pixel-wise height of US images. The origin of frame $\{us\}$ was set at the upper left of the B-mode image, while the origin of frame $\{p\}$ was set in the middle of the probe tip (Fig.~\ref{Fig_coordinate_system}). Thus, $^{p}_{us}\textbf{T}$ is calculated as follows:

\begin{equation}\label{eq_I_to_probe}
^{p}_{us}\textbf{T} = \begin{bmatrix}
0 & 0 & -1 & 0\\
-\frac{L_p}{W_{us}} & 0 & 0 & \frac{L_p}{2}\\
0 & \frac{D}{H_{us}} & 0 & \varepsilon_0\\
0 & 0 & 0 & 1\\
\end{bmatrix}
\end{equation}
where $L_p=37.5~mm$, $D = 55~mm$, and hyper parameter $\varepsilon_0$ is used to neutralize the indeterminacy of US elements configuration.

\par
In addition, regarding the hand-eye calibration, similar implementation procedures are carried out as~\cite{jiang2021motion}. To compute the transformation matrix $^{b}_{c}\textbf{T}$ between the base frame $\{b\}$ and the camera frame $\{c\}$, paired coordinate descriptions of the points (intersection points on two chessboards) in frame $\{c\}$ and frame $\{b\}$ are recorded. The coordinate representations in $\{c\}$ are computed using OpenCV while the representations in $\{b\}$ are obtained by manually moving a custom-designed pointer tool to the same intersection points. More implementation details can be found in~\cite{jiang2021motion}.

\subsection{Compliant Control Architecture}
\par
Since the acoustic impedance of air is much lower than human tissues (usually over $1000$ times lower), the reflection coefficient at the air-skin interface is around one (i.e., almost all signal gets reflected). Thus, a certain contact force between a US probe and object surface is necessary to guarantee the visibility of target anatomies. To eliminate the negative influence of non-homogeneous deformations in 3D results, a compliant controller using the built-in joint torque sensors was employed~\cite{hennersperger2016towards, jiang2021autonomous}. The Cartesian compliant controller law is described as follows:

\begin{equation}\label{eq_impedance_law}
\tau = J^{T}[K_m(x_d - x_c) + F_d] + D(d_m) + f_{dyn}(q,\Dot{q}, \Ddot{q})]
\end{equation}
where \revision{$\tau \in R^{7\times 1}$} is the target torque for all joints, \revision{$J^{T}\in R^{7\times 6}$} is the transposed Jacobian matrix, \revision{$x_d \in R^{6\times 1}$} and \revision{$F_d \in R^{6\times 1}$} are the desired Cartesian position and force, \revision{$x_c\in R^{6\times 1}$} is current position, \revision{$K_m \in R^{6\times 6}$} represents the Cartesian stiffness, \revision{$D(d_m)\in R^{7\times 1}$} is the damping term, and \revision{$f_{dyn}(q,\Dot{q}, \Ddot{q})\in R^{7\times 1}$} is the dynamic model of the robotic arm. 

\par
Considering the characteristic of US examinations, the constant force is maintained in the direction of the probe centerline by a 1-DoF compliant controller, while the other 5-DoF motion is controlled to accurately follow the planned sweep trajectory. These two types of controllers are fused by assigning varied stiffness values. High stiffness (translational DoF: $2000~N/m$ and rotational DoF: $200~Nm/m$) are assigned to the DoFs controlled in position mode while the stiffness of the compliant DoF is set between $[125, 500]~N/m$ for different human tissues~\cite{hennersperger2016towards}.

\section{Motion-Aware Robotic US Scanning}
\par
This section describes the characteristic aspects of the proposed system. The method used to automatically create scan trajectory from a template image is described in Section III~A. In Section III~B, a confidence-based in-plane orientation optimization is proposed to optimize the contact condition between probe and object surface. Section III~C demonstrates a markerless motion compensation method designed based on surface point cloud registration and a fine adjustment for further eliminating the stitching gap in 3D view. 

\subsection{Surface Registration and Trajectory Transfer}
\par
Since the object position varies in different examinations, a depth camera is employed to gain information about the current environmental setup. To properly execute US scan for the target artery, the annotated artery is required to be transferred into camera frame $\{c\}$. To obtain such transformation $^{c}_{ct}\textbf{T}$, the iterative closest point (ICP) algorithm is used to register the surface point cloud $\textbf{P}_{ct}$ obtained from a pre-operative CT image to the live point cloud $\textbf{P}_{c}$ computed from the RGB-D camera. ICP proved to be sufficient for the targeted task in preliminary experiments. To achieved uniformly distributed $\textbf{P}_{ct}$, the Poisson disc sampling algorithm~\cite{dunbar2006spatial} is employed instead of the random sampling approach. 
Since ICP is sensitive to the initial result, the multiscale feature resistance approach is used to extract unique features from both point clouds for computing the initial transformation~\cite{holz2015registration}. Afterward, the iterative algorithm is used to accurately align $\textbf{P}_{ct}$ to $\textbf{P}_{c}$.

\par
\revision{To correctly and robustly align $\textbf{P}_{ct}$ to $\textbf{P}_{c}$, the distinctiveness of each feature descriptor is important, which will significantly affect the number of possible ambiguous correspondences between two point clouds. To this end, the multiscale feature resistance approach is employed, which computes the same class of feature descriptor at various scales and further refines the point pairs that were proved to be distinctive over multiple scales. The default scales are $0.5$, $1.0$ and $1.5$. Then, the paired correspondences are further filtered in terms of consistency based on the distance threshold, which is empirically selected as $0.25$ based on the final performance. The implementation is based on the Point Cloud Library (PCL)\footnote{https://pointclouds.org/} and more details can refer~\cite{holz2015registration}.
}

\par
Since the artery is located inside the arm, the artery trajectory is required to be projected to the surface for generating an executable scanning trajectory. To alleviate potential negative influences of registration error between $\textbf{P}_{ct}$ and $\textbf{P}_{c}$ on creating scanning trajectory, the point cloud of the artery surface in CT $\textbf{P}_{ct}^{art}$ is directly transferred into the camera frame. Then, $\textbf{P}_{ct}^{art}$ can be exactly projected onto the object surface captured by the camera. Based on the hand-eye calibration result described in Section~II~B, the scanning trajectory is computed.

\subsubsection{Artery Centerline Extraction} 
\par
The artery surface point clouds have a tubular structure. Besides, regarding the geometry of limb arteries, the length along the vascular centerline is significantly longer than in other directions (namely, artery diameter). To extract the vascular centerline for generating scanning trajectory, the principal component analysis (PCA) is employed to define a new orthogonal coordinate system for optimally describing variance of the data. The first principal component is the direction making the projections achieve the largest variance, namely the eigenvector corresponding to the largest eigenvalue of the artery point cloud's covariance matrix. To estimate the vascular centerline, the average value of all points inside a small artery segmentation is computed to an approximate local center point. A small distance interval $d_{in}$ is used to generate multiple center points between the minimal and maximal value of the projections in the first principal component. The vascular centerline is approximated by connecting the center points successively.

\subsubsection{Scanning Trajectory on Object Surface} 
\par
After deriving the vascular centerline, an executable scanning trajectory on the object surface should be further generated. To avoid potential safety issues caused by improper scanning trajectories, e.g., inside of the arm, the scanning trajectory is generated as following steps. Since the robot and the experimental table are parallel to the ground, a continuous trajectory can be created by projecting the estimated vascular centerline onto the surface in the direction of $Z_{b}$. To achieve this, the center points of the artery $(x, y, z)$ computed in last section is replaced with $(x, y, z^{arm}_{max})$, where $z^{arm}_{max}$ is the maximum value of the arm surface point cloud $\textbf{P}_{ct}$ in $Z_{b}$ direction. Then $K$-nearest neighbors (KNN) approach is used to search for the $K_{st}=5$ nearest points around $(x, y, z^{arm}_{max})$, individually, in terms of Euclidean distance. Then, the average of these $K_{st}$ neighbors is used as key points of the scanning trajectory for RUSS. 

\par
\begin{figure}[ht!]
\centering
\includegraphics[width=0.40\textwidth]{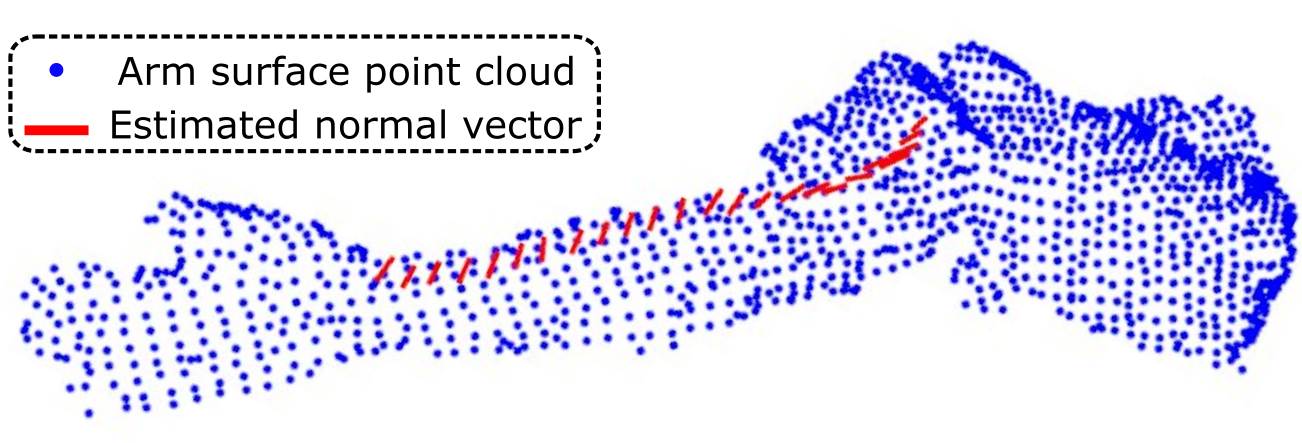}
\caption{Computed scanning trajectory on the arm surface. The red line represents the estimated normal direction at individual points on the trajectory. 
}
\label{Fig_sweep_trajectory}
\end{figure}

To fully control the probe during the scanning, the probe orientation needs to be further determined. To improve the contrast of the resulting B-mode imaging, Ihnatsenka~\emph{et al.} and Jiang~\emph{et al.} suggested that the probe should be aligned in the normal direction of the contact surface $\Vec{n}_i$~\cite{ihnatsenka2010ultrasound,jiang2020automatic}. Here, the normal direction is quickly estimated using computer vision techniques based on the neighboured points around the key points. Compared with the force-based method~\cite{jiang2020automatic}, the vision-based method is able to quickly compute all the normal directions along the whole trajectory rather than only the current contact point. This process is implemented using PCL. The estimated normal directions at various positions are shown in Fig.~\ref{Fig_sweep_trajectory}. To properly visualize the target vessel, the probe centerline is aligned to the computed normal direction and the long axis of the probe ($Y_{p}$) is aligned to be orthogonal to the scanning path.

\subsection{Confidence-based Probe Orientation Correction}
\par
\begin{figure}[ht!]
\centering
\includegraphics[width=0.45\textwidth]{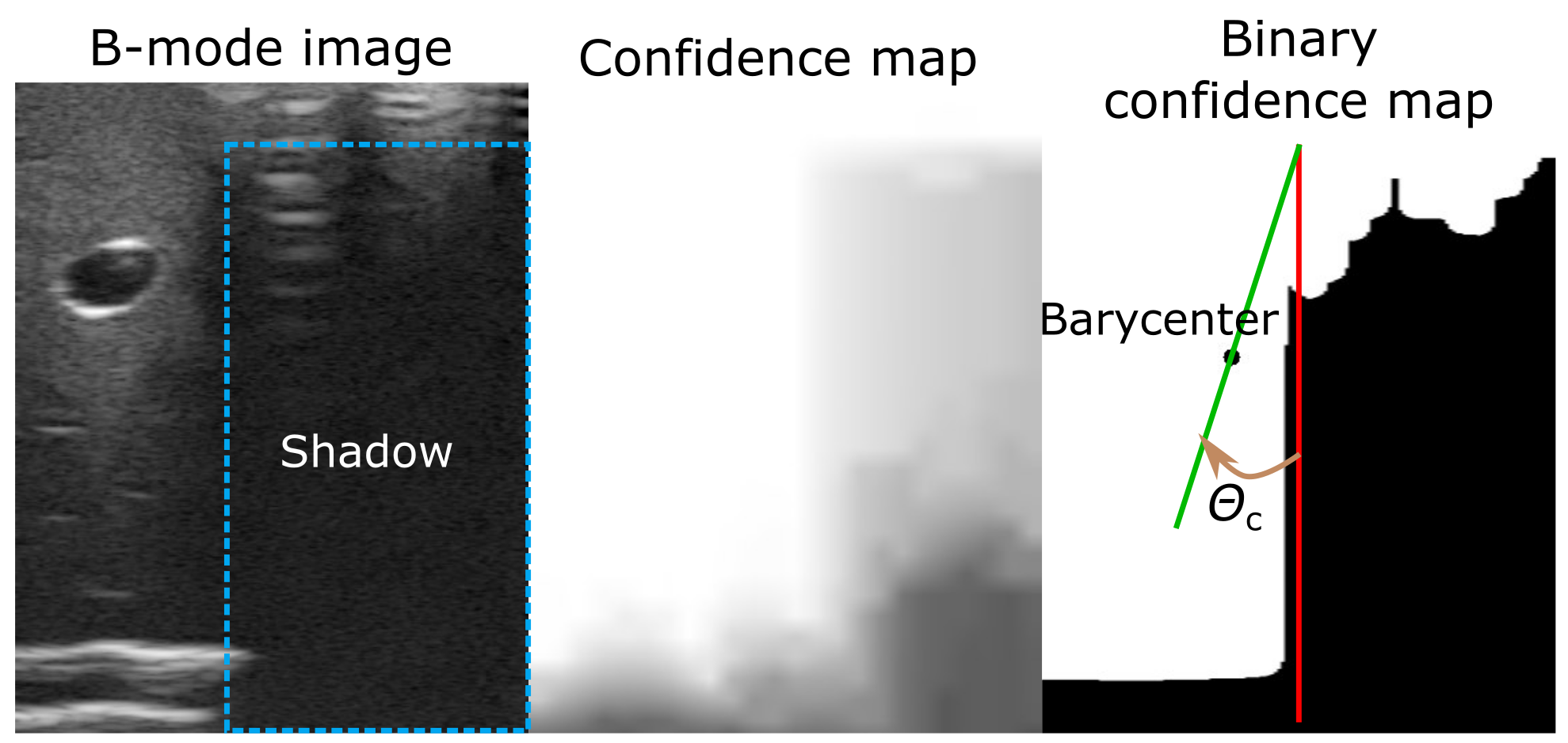}
\caption{Confidence-based orientation correction. The left, middle and right images are B-mode image, computed confidence image, and binary confidence map. In this case, computed correction angle is $\theta_c = 10.4^{\circ}$.
}
\label{Fig_confidence_correction}
\end{figure}

US confidence map was originally introduced by Karamalis~\emph{et al.}~\cite{karamalis2012ultrasound}. They provide a pixel-wise metric to assess the image quality by computing the loss of emitted US signals by the transducer. Regarding the computed confidence maps of B-mode images, a probabilistic map \revision{$C\in {{\mathbb {R}}^{2}}\xrightarrow{}[0, 1]$} is created~(see Fig.~\ref{Fig_confidence_correction}). The top white pixels ($1$) represent the strongest signal and the bottom black pixels ($0$) represent no US signal arrived. US confidence map has been employed for different aims, e.g., optimizing contact force~\cite{virga2016automatic} and tracking target anatomy~\cite{chatelain2017confidence}.

\par
In this study, the computed scanning trajectory may not be perfect for scanning due to the error of both hand-eye calibration and surface registration (ICP). The use of a human-like arm phantom with an uneven surface aggravates the negative influence of such errors. More specifically, the US probe may not be able to fully contact the arm surface, which will result in shadow in the imaging (see Fig.~\ref{Fig_confidence_correction}). To correctly display the anatomy under the shadow, a confidence-based orientation correction approach is proposed. To this end, the confidence map is further transferred into a binary image by applying a probabilistic threshold ($T_{com}$). This emphasizes the difference between the well-contacted part and the non-contact part in the resulting images. Then, the weighted barycenter point \revision{$\zeta_{c}\in R^{2\times1}$} can be calculated as follows:

\begin{equation} \label{eq_Barycenter}
\zeta_{c} = \frac{1}{\aleph_{c}} \sum_1^{H_{us}} \sum_1^{W_{us}} C(h, w)\left[ h, w \right]^T
\end{equation}
where \revision{$C(h, w)$ is binary confidence value at pixel grid $(h, w)$, $H_{us}= 550$ and $W_{us}=375$ are the image's height and width in pixel.} $\aleph_{c} = \sum_{(h, w)\in \Omega} C(h, w)$ is the accumulated confidence value over the entire image ($\Omega$). 

\par
Then, the in-plane rotational adjustment angle $\theta_c$ can be calculated by connecting the barycenter and the top center point of image view (see Fig.~\ref{Fig_confidence_correction}). To ensure that the probe is fully in contact with the object surface, multiple in-plane orientation corrections may be required at one position. To reduce the time consumed at each position, the pose (\revision{$R_p \in R^{3\times 3}$}) of the remaining $N_{up}$ points in the trajectory are updated as follows:

\begin{equation}~\label{eq_update_trajectory}
R_p^{'}(i_c+i) = R_p(i_c+i) R_{x}(-\eta_i\theta_c)
\end{equation}
where $i_c$ represents the iteration of current position in the trajectory, $i=1, 2,.., N_{up}$, \revision{$R_{x}\in R^{3\times 3}$} is the rotation matrix around $X$ direction of probe coordinate system, $\eta_i = \frac{d_{N_{up}-i+1}^2}{\sum_{j=1}^{N_{up}}d_j^2}$ is the weight coefficient used to update the next $N_{up}$ positions.

\subsection{Movement Identification and Compensation}
\subsubsection{Movement Monitor System}
\par
To effectively detect the potential movement (expected or unexpected) during the scanning, the UNet-VGG16 architecture~\cite{balakrishna2018automatic, pravitasari2020unet} is employed to segment the arm from RGB images. The U-Net was proposed by Ronneberger~\cite{ronneberger2015u} for segmentation tasks based on Fully convolutional networks (FCNs). \revision{U-Net consists of an encoder and decoder. To achieve accurate segmentation, skip connections are used to transfer the features from the encoder to the decoder. This is one of the most common network structures used for semantic segmentation. Compared to training the UNet encoder from scratch, VGG16~\cite{simonyan2014very} is already trained on a large-scale ImageNet classification dataset (over 14 million images)~\cite{deng2009imagenet}, which has already learned enough features. Thereby, UNet-VGG16 replaces the UNet encoder with a well-trained VGG16 model to extract the image features, which can improve the overall performance of the network and as well as reduce the training time and training dataset size~\cite{pravitasari2020unet}. 
}


\par
To train an UNet-VGG16 model to efficiently segment the arm in this setup, $270$ images were collected by changing both the camera position and the environments, e.g., different backgrounds, occlusion caused by the robotic arm and US probe, and with/without humans in the field of view. Then, the arm was manually annotated on all images using LabelMe software~\cite{russell2008labelme}. To augment the data for better generalization, the images are randomly flipped and the color are also randomly modified by changing the related parameters, i.e., brightness and contrast. Finally, $3240$ images are generated. \revision{The ratio between the training and validation data sets is $8:2$. The size of the input images is $512\times512$.} In addition, the pixel-wise mean squared error $L_{mse}$ is employed as the loss function to train the network. 

\begin{equation}\label{eq_loss_cross_entropy}
L_{mse} = \frac{1}{N}\sum_{c=1}^{C_I} \sum_{i=1}^{H_c} \sum_{j=1}^{W_c}{[y_c(i,j) - \hat{y}_c(i,j)]^2}
\end{equation}
where \revision{$N=C_I\cdot H_c \cdot W_c$; $C_I$, $H_c$, $W_c$ are the number of channels, height and width, respectively.} $y$ and $\hat{y}$ are the annotated image and output of the network. 

\par
To validate whether the proposed pipeline is compatible for real scenarios, the second UNet-VGG16 model was trained separately for human volunteers. To avoid articulated movements, only the forearm was selected as target object. The dataset of human arm was generated using the same manner as arm phantom data. Finally, $3560$ labeled images from two healthy volunteers (BMI: $25.0\pm2.8$, age: $28.5\pm1.5$) were separated as $8:2$ for training and validation.

\par
After obtaining the binary masks using the trained model, the dice coefficient $C_d$ between the segmented results from the current frame ($i$-th) and the previous frame ($i-j$-th) is calculated as follows: 

\begin{equation}\label{eq_dice}
C_d = \frac{2~|S_{i-j}\cap S_{i}|}{|S_{i-j}|+|S_{i}|}
\end{equation}
where $S_{i}$ is the segmented binary results of $i$-th camera image. 

\subsubsection{Movement Compensation}
\par
If a movement happened during the scanning, the real-time dice coefficient decreases correspondingly. Thus, $C_d$ is used as the metric to identify the object movement. Once  $C_d$ becomes smaller than a preset threshold $T_{dice}$, a signal is emitted to inform the controller to stop scanning via ROS. Based on the experimental results, $T_{dice}$ was empirically set as $0.95$ to guarantee both the sensitivity and the robustness of motion detection results in our setup. To properly resume the sweep from the breaking point, the transformation between previous (before the movement) and current objects is calculated using ICP (see Fig.~\ref{Fig_workflow}). Here, the point clouds of the arm surface are created based on the mask of RGB images. After properly assigning the depth information to the pixels located inside the bounding box of the mask, a raw point cloud can be generated. To extract the arm, the plane segmentation algorithm is applied to remove the points under the table surface and an additional depth filter is employed to remove the points higher than a certain value in $Z_b$ direction of frame $\{b\}$. To validate whether the registration results are good enough for continuing the scanning from the break point, an ArUco marker is rigidly placed on the table. To quantitatively access the registration results, the motion compensation error $e_{mc}$ is computed as follows:

\begin{equation}\label{eq_movement_registration_error}
e_{mc}=\left \|\mathbf{P}_{ar}^{'}-(\mathbf{R}_{mc} \cdot \mathbf{P}_{ar} + \mathbf{V}_{mc})\right\|
\end{equation}
where \revision{$\mathbf{P}_{ar}\in R^{3\times1}$} and \revision{$\mathbf{P}_{ar}^{'}\in R^{3\times1}$} are the positions of the ArUco marker before and after the movement, respectively, while  \revision{$\mathbf{R}_{mc}\in R^{3\times3}$} and \revision{$\mathbf{V}_{mc}\in R^{3\times1}$} are the computed rotation matrix and translation vector.

\par
\revision{
The RUSS automatically resumes the scanning from the breaking point if $e_{mc}$ is small enough ($<1~cm$). Otherwise, the RUSS automatically ends the sweep. Compared with the state-of-the-art marker-based approach~\cite{jiang2021motion}, only camera images make it more suitable for real clinical routines without the need to carefully configure the markers on various patients. }

\subsubsection{Fine Adjustment Algorithm}
\par
To well stitch the two or multiple sweeps when movement happens during scans, \revision{a two-step fine adjustment procedure is further developed to displaying a complete and continuous 3D anatomy.} To achieve this, the probe pose of the last frame in the first sweep (before movement) (\revision{$\textbf{T}_{be}^{la} \in R^{4\times 4}$}) and the probe pose of the first frame in the second sweep (after movement) (\revision{$\textbf{T}_{af}^{fi} \in R^{4\times 4}$}) are used. To overlap these two frames, the tracked probe pose (\revision{$\textbf{T}_{be} \in R^{4\times 4}$}) of the first sweep can be updated as follows:

\begin{equation}\label{eq_fine_tuning_first}
\begin{bmatrix}
\textbf{R}^{'}_{be}(i) & \textbf{P}^{'}_{be}(i) \\
0 & 1\\
\end{bmatrix}
=\left[ (\textbf{T}_{af}^{fi})^{-1} \textbf{T}_{mc} \textbf{T}_{be}^{la}\right]^{-1} \textbf{T}_{mc} \textbf{T}_{be}(i)
\end{equation}
where $\textbf{T}_{mc}$ is the homogeneous expression of $\mathbf{R}_{mc}$ and $\textbf{V}_{mc}$ in Eq.~(\ref{eq_movement_registration_error}).

\par
\revision{To further achieve good continuity of vascular boundary in 3D, an in-plane translational adjustment is carried out. The translational adjustment is computed based on} the two centroids ($\textbf{P}^c_{be}$ and $\textbf{P}^c_{af}$) of the vessel of interest from the last frame in the first sweep and the first frame in the second sweep, respectively. The centroid is computed based on the binary mask computed using a well-trained U-Net model as~\cite{jiang2021autonomous}. To train the network, $3369$ B-mode images were used. \revision{Since the batch size was set to ten to improve the generalisability, the batch normalization layers used in the original U-Net~\cite{ronneberger2015u} are replaced by the group normalization approach. The training details are exactly the same as~\cite{jiang2021autonomous}: the ratio between training and validation data set is $9:1$, the learning rate is initialized as $0.001$, the optimization method is ADAM optimizer, and the epoch is $10$.} Finally, the tracked probe pose of the sweep obtained before the movement is further updated as follows:

\begin{equation}\label{eq_fine_tuning_second}
\textbf{P}^{''}_{be}(i) = \textbf{P}^{'}_{be}(i) + ^{b}_{us}\textbf{R} (\textbf{P}_{af}^c - \textbf{P}_{be}^c)
\end{equation}

\section{Results}

\subsection{Segmentation Results on Phantoms and Human Arm}
\begin{figure}[ht!]
\centering
\includegraphics[width=0.40\textwidth]{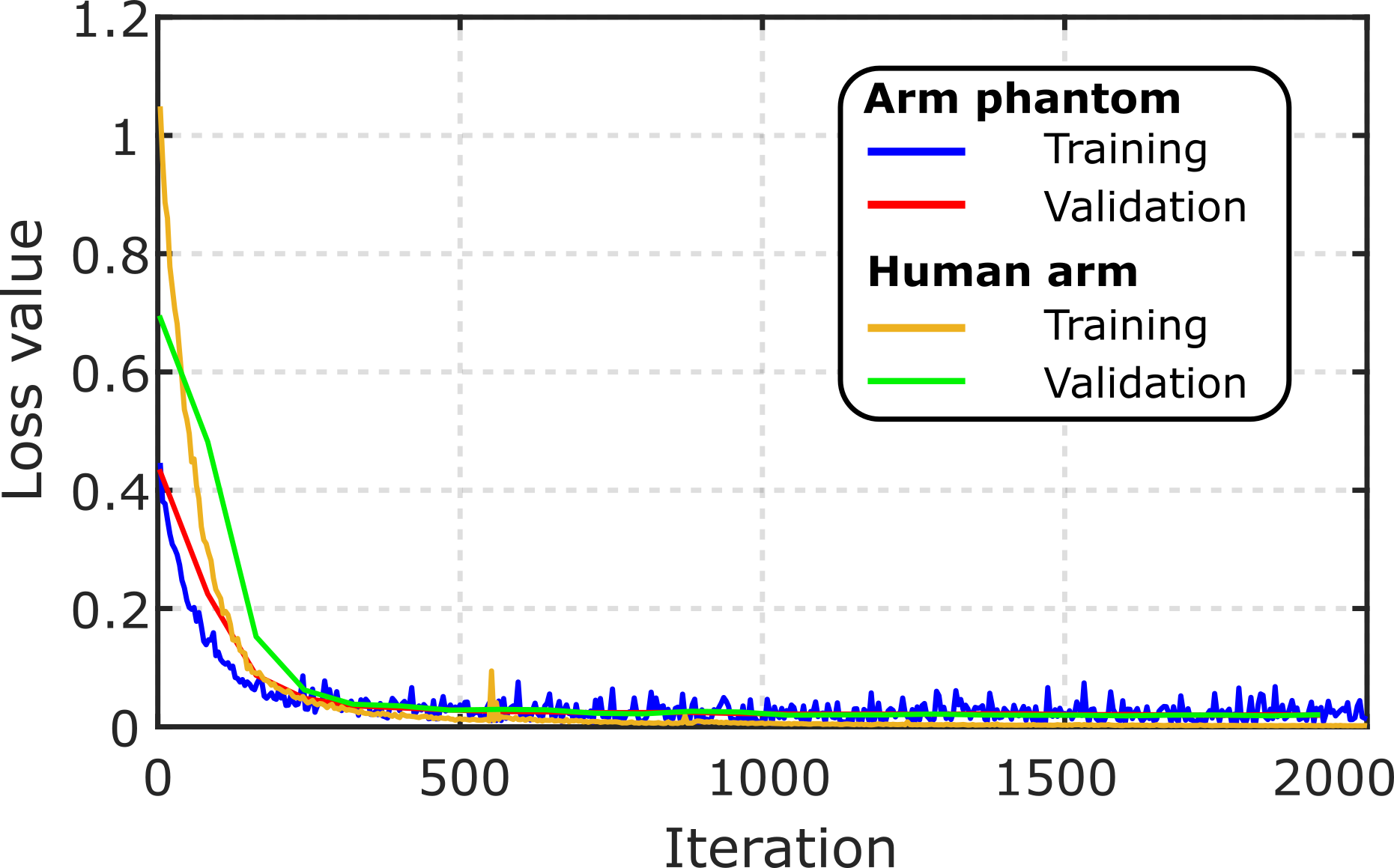}
\caption{\revision{The training loss and validation loss of the UNet-VGG16 for arm surface segmentation from RGB images.}
}
\label{Fig_loss}
\end{figure}

\revision{The UNet-VGG16 network parameters were optimized using Adam~\cite{kingma2014adam} on a single GPU (Nvidia GeForce GTX 1080). The learning rate was set to $10^{-3}$ at the beginning and reduced by a factor of ten when the loss changes were less than 0.0001 for ten subsequent steps. The batch size and the training epochs are $4$ and $10$, respectively.} \revision{It can be seen from Fig.~\ref{Fig_loss} that both training loss and validation loss were quickly reduced at the beginning and
gradually converged after $500$ iterations.} The performance of the well-trained models on unseen images of arm phantom and human forearms are summarized in TABLE~\ref{Table_segmentation_result}. The average dice coefficient achieved $0.94$ and $0.95$ on arm phantom and human forearms, respectively. \revision{Besides, the segmentation time for each image only takes around $5~ms$.}

\begin{table}[!ht]
\centering
\caption{Performance of Segmentation Algorithm}
\label{Table_segmentation_result}
\begin{tabular}{ccccc}
\noalign{\hrule height 1.2pt}
Dataset   & Dice Coefficient & Time (ms) & Samples \\ 
\noalign{\hrule height 1.0 pt}
Arm phantom   & $0.94\pm0.04$ & $5.3\pm0.6$ & $633$\\

Volunteers forearm   & $ 0.95\pm0.03 $ & $5.4\pm0.6$ & $704$\\

\noalign{\hrule height 1.2 pt}
\end{tabular}
\end{table}

\par
\begin{figure}[ht!]
\centering
\includegraphics[width=0.49\textwidth]{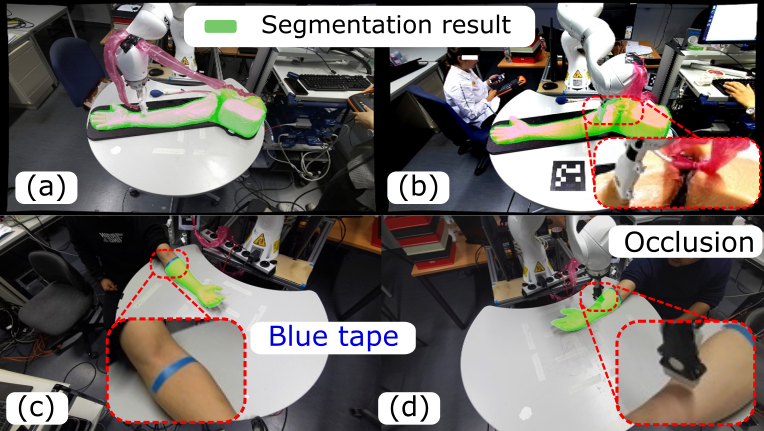}
\caption{Segmentation performance on unseen images. (a) and (b) are results of unseen phantom images without and with occlusion, (c) and (d) are the results of unseen images from two different volunteers. The computed dice coefficients for (a), (b), (c) and (d) are $0.96$, $0.95$, $0.97$ and $0.96$, respectively. 
}
\label{Fig_segmented_results}
\end{figure}

To demonstrate the performance of the segmentation result, four representative results on arm phantom and two different human arm are shown in Fig.~\ref{Fig_segmented_results}. Regarding the arm phantom, Fig.~\ref{Fig_segmented_results} (a) and (b) display segmentation results form RGB images when the arm surface is without/with occlusion caused by US probe, respectively. Regarding the experiments on volunteers, a normal blue tape was warped around human elbow. The tape servers as boundary of forearm and its position is easily changed and determined based on target anatomies by human operators. The segmentation results on two different volunteers are shown in Fig.~\ref{Fig_segmented_results} (c) and (d). Even when the occlusion happens, the forearm boundary is still completely extracted. These results demonstrate that well-trained models has the potential to accurately extract the target objects. The dice coefficient of these four representative results are $0.96$, $0.95$, $0.97$ and $0.96$, respectively. \revision{Accurate and robust segmentation results have been achieved on both phantom and human arms, which demonstrates that the position of the RGB-D camera will not significantly affect the results.} In addition, more live segmentation results can be found in the attached video.

\subsection{ICP-Based Surface Registration Performance}
\par
\begin{figure}[ht!]
\centering
\includegraphics[width=0.48\textwidth]{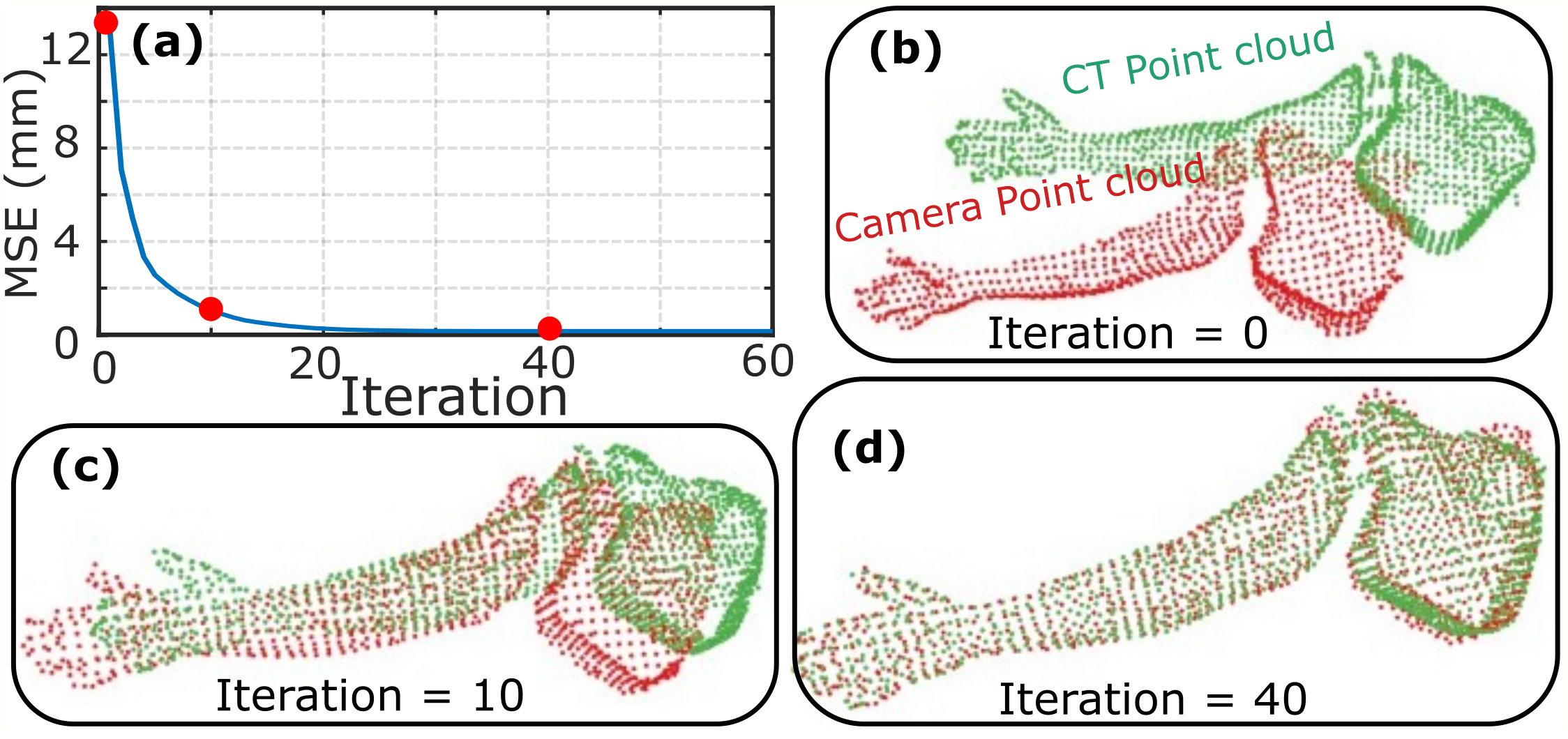}
\caption{Surface registration results. (a) is the mean squared error (MSE) of distance between corresponding points in the two point clouds, (b), (c) and (d) are the results when the iteration is $0$, $10$ and $40$, respectively. The sizes of preoperative $\textbf{P}_{ct}$ and camera-based $\textbf{P}_{c}$ are $1379$ and $925$, respectively. 
}
\label{Fig_fitness_results}
\end{figure}

\revision{To demonstrate the performance of the ICP-Based surface registration approach in our setup, some representative results have been shown in Fig.~\ref{Fig_fitness_results}. The mean squared error (MSE) of the distance between corresponding points in the two point clouds ($\textbf{P}_{ct}$ and $\textbf{P}_{c}$) quickly decreased at the beginning and converged ($0.2~mm$) after $25$ iterations. The initial alignment is shown in Fig.~\ref{Fig_fitness_results}~(b), which is computed using the multiscale feature resistance approach~\cite{holz2015registration}. Then the result after $10$ and $40$ iterations are depicted in Fig.~\ref{Fig_fitness_results}~(c) and (d). It can be seen that the point cloud obtained using an RGB-D camera can be well registered to a preoperative template (MSE is $0.16~mm$).}

\par
\begin{figure}[ht!]
\centering
\includegraphics[width=0.48\textwidth]{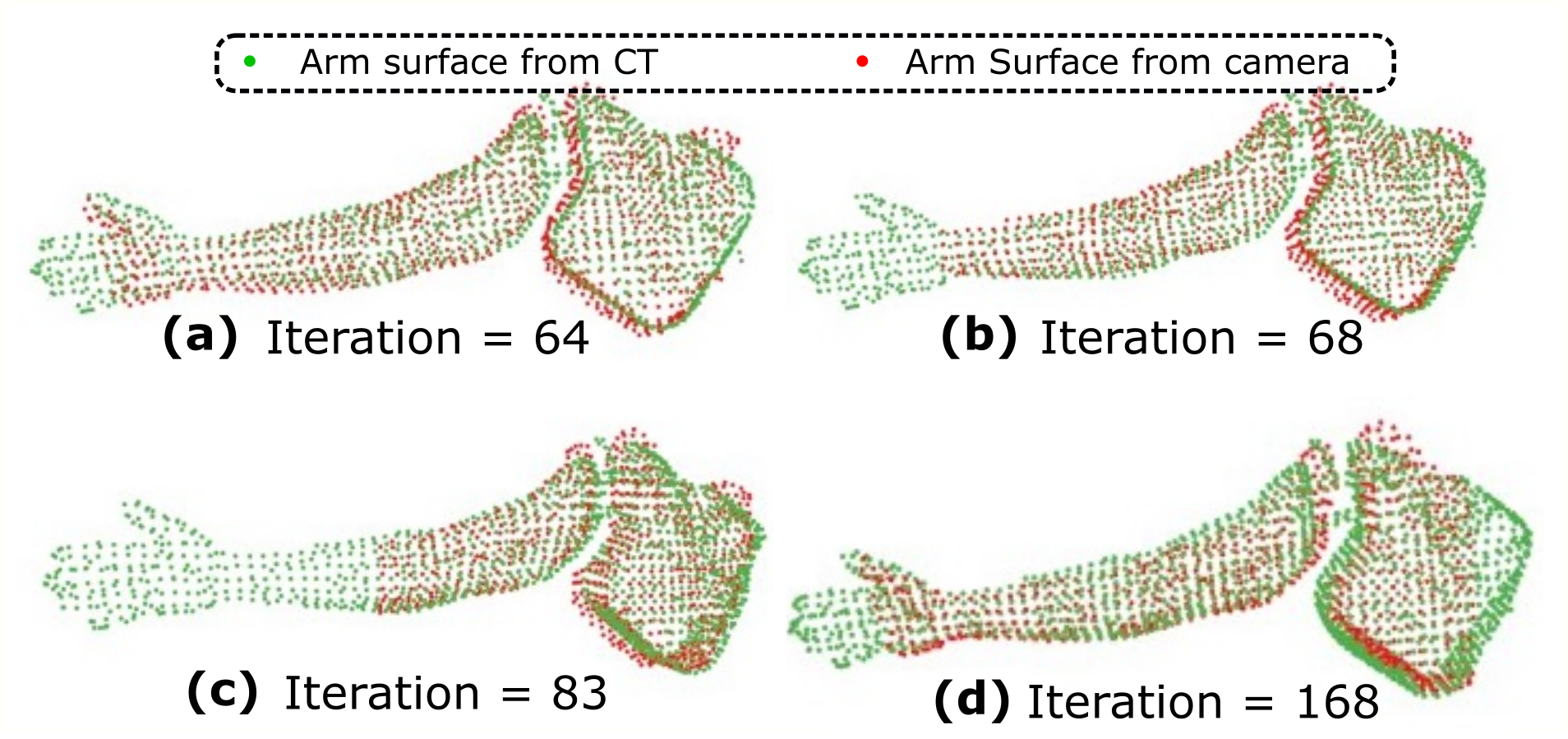}
\caption{Surface registration results between preoperative point cloud and incomplete camera point cloud. In (a), (b), and (c), the camera point cloud $\textbf{P}_{c}$ is cropped from $10\%$, $20\%$ and $40\%$, respectively, on the first principal direction. In (d), $\textbf{P}_{c}$ is cropped by two planes at $10\%$ and $90\%$ on the first principal direction. 
}
\label{Fig_crop_results}
\end{figure}

Considering the potential occlusion in a real scenario, the extracted point cloud from an RGB-D camera $\textbf{P}_{c}$ could be incomplete. To further validate the robustness of the ICP-based Registration, $\textbf{P}_{c}$ is cropped by a plane being orthogonal to the first principal direction computed using PCA. To investigate the influences caused by various levels of occlusion, $\textbf{P}_{c}$ is cropped by the plane at $10\%$, $20\%$ and $40\%$ on the first principal direction [Fig.~\ref{Fig_crop_results} (a), (b) and (c), respectively] and two planes at $10\%$ and $90\%$ [Fig.~\ref{Fig_crop_results} (d)]. In all these four setups, good results (Fig.~\ref{Fig_crop_results}) are achieved after a certain iterations ($64$, $68$, $83$, and $168$, respectively). The final MSE are $0.3$, $1.6$, $4.4$ and $2.7~mm$, respectively. \revision{The iterative registration process takes around $336~ms$ on average. }


\begin{figure*}[ht!]
\centering
\includegraphics[width=0.80\textwidth]{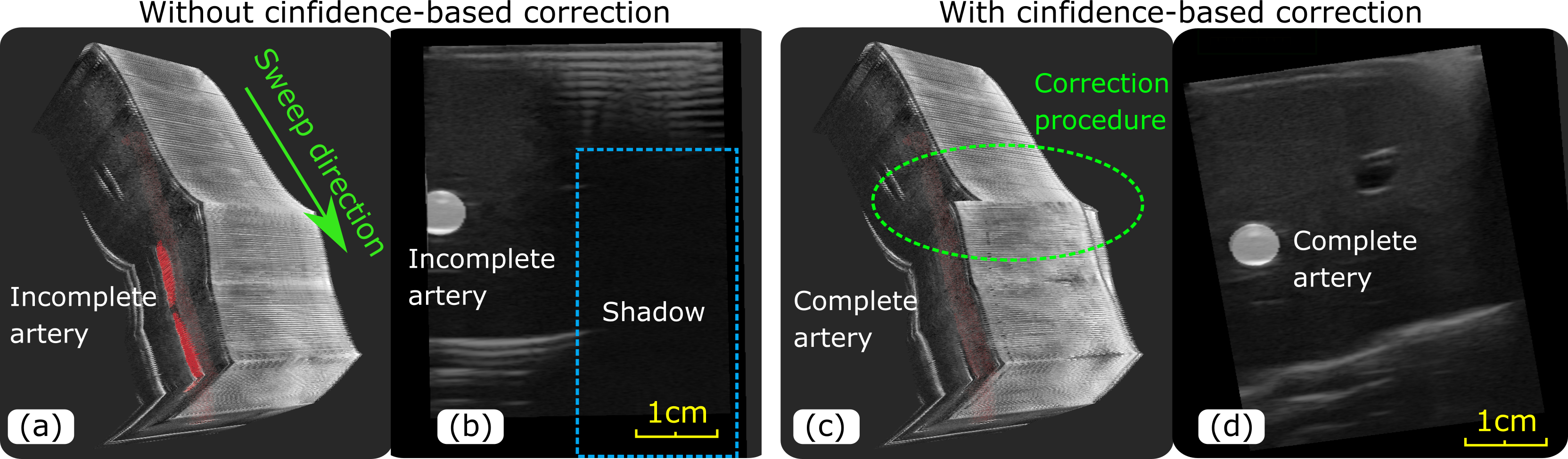}
\caption{Performance of the confidence-based orientation correction algorithm. (a) and (c) are the 3D results obtained without and with confidence-based correction. (b) is the B-mode images where the target artery geometry is partly out of the view, while (d) is the corrected result where the artery is completely displayed in the imaging view.
}
\label{Fig_confidence_correction_3d}
\end{figure*}

\subsection{Performance of Confidence-based Trajectory Correction}
Due to the error of hand-eye calibration and CT-to-object registration, the computed trajectory may result in non-optimal contact during the scan. This reduces the quality of US imaging, like introducing a shadow in the image view as shown in Fig.~\ref{Fig_confidence_correction_3d}~(b). Thus, a confidence-based orientation correction is developed to guarantee good contact condition between the probe and the arm surface during the scan. To demonstrate the performance of the confidence-based correction, two sweeps with an exactly same initial trajectory are performed. The representing results are depicted in Fig.~\ref{Fig_confidence_correction_3d}. 

\par
It can be seen from Fig.~\ref{Fig_confidence_correction_3d}, part of the target artery is out of the imaging view for the case without correction. This is represented by the red part in 3D and white circle (incomplete) in 2D [Fig.~\ref{Fig_confidence_correction_3d}~(a) and (b)], respectively. In the case involving correction, once a shadow is detected using the confidence map, an in-plane adjustment angle $\theta_c$ is computed and the correction is automatically carried out by the robot. After such correction, the entire artery geometry has been successfully visualized in the imaging view [see Fig.~\ref{Fig_confidence_correction_3d}~(d)]. In addition, since the poses of the remaining positions in the trajectory are also updated using Eq.~(\ref{eq_update_trajectory}), the target vessel is completely visualized inside the imaging view [see Fig.~\ref{Fig_confidence_correction_3d}~(d)]. This result demonstrates that the proposed confidence-based orientation can help to improve the visibility of the target object during the sweep.

\subsection{Movement Compensation Results}
\par
The compensation is done by two successive procedures, 1) surface registration-based compensation and 2) the fine adjustment process using both robotic tracking information and corresponding B-mode images. To qualitatively and quantitatively validate the compensation algorithm, a complete artery of an arm phantom is used as the target object (total length is around $440~mm$).  

\subsubsection{Performance of Surface Registration-based Compensation}
\par
To quantitatively assess the surface registration performance, the phantom arm together with the used table was randomly moved inside of a rectangle ($110\time140mm$) and the maximum rotation angle variation was $80^{\circ}$. To assess the performance of the surface registration-based compensation, an ArUco marker is physically placed on the flat table (see Fig.~\ref{Fig_coordinate_system}). The real-time pose (position and orientation) of the ArUco marker can be computed using a ROS package\footnote{https://github.com/pal-robotics/aruco\_ros}. Thus, the motion compensation error $e_{mc}$ is computed by Eq.~(\ref{eq_movement_registration_error}). 

\par
\begin{figure}[ht!]
\centering
\includegraphics[width=0.38\textwidth]{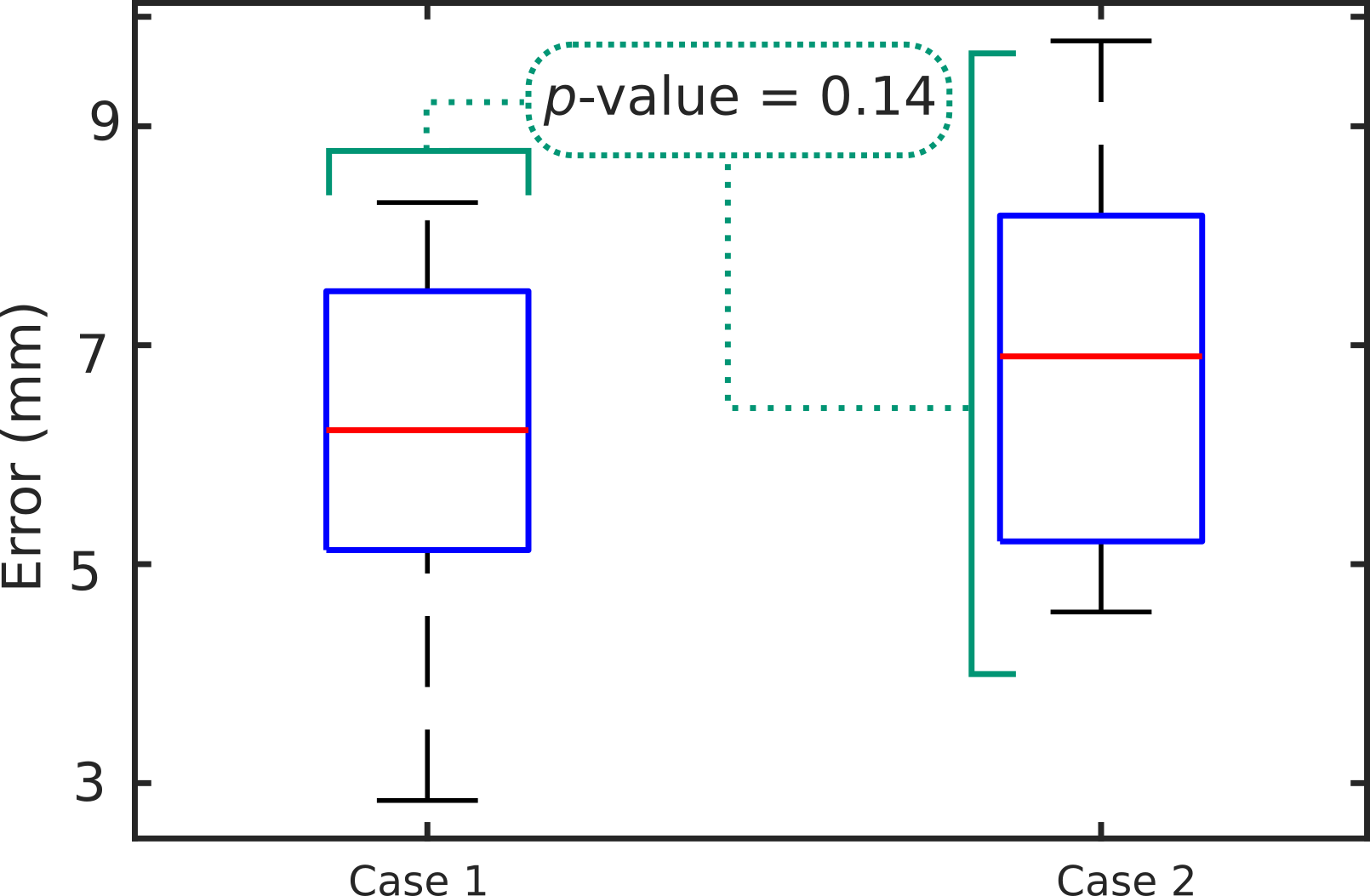}
\caption{Performance of the surface registration-based compensation algorithm. Case 1 and case 2 represent the situations where the phantom arm surface is not and is occluded by the robotic manipulator, respectively.}
\label{Fig_compensation_error}
\end{figure}

Since the arm surface could be partly occluded by the used robotic manipulator [Fig.~\ref{Fig_segmented_results}~(b) and (d)] during the scanning, the experiments are separately performed when (case 1) the robotic arm is out of the view and (case 2) is in contact with the phantom arm surface. The experiments were repeated $20$ times for each case. The final results are shown in Fig~\ref{Fig_compensation_error}. The $e_{mc}$ ($\pm$SD)  of case 1 and case 2 are $6.1\pm1.6~mm$ and $6.9\pm1.6~mm$, respectively. The results obtained when there is an occlusion in the camera view (case 2) is slightly larger than the results of the ideal case (case 1). Besides, based on a t-test (probability $p=0.14>0.05$), it was concluded that no significant difference exists between the two experimental sets.
\secrevision{To perform the t-test, the results obtained from both case 1 and case 2 are proved to be normally distributed using the Lilliefors goodness-of-fit test.} Considering our application, since $6~mm$ is much smaller than the probe width $37.5~mm$, the surface registration-based compensation algorithm is promising to be used for compensating a large movement that happens during the scanning. Thereby, a complete geometry of the anatomy can be computed as in Fig.~\ref{Fig_3D_compesated}~(b).

\subsubsection{Performance of Fine Tuning Procedure}
\par
\begin{figure}[ht!]
\centering
\includegraphics[width=0.48\textwidth]{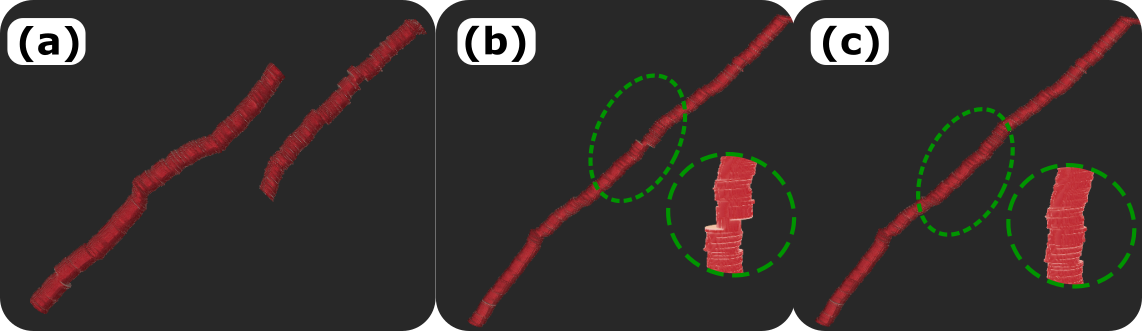}
\caption{Performance of compensation method. 3D images of a vessel (a) without any compensation, (b) with surface registration-based compensation, and (c) further with a step-wise fine adjustment procedure.
}
\label{Fig_3D_compesated}
\end{figure}

After surface registration-based compensation, two partitions have been stitched together. However, due to the error $e_{mc}$, there is a significant gap and dislocation at the connected part as in Fig.~\ref{Fig_3D_compesated}~(b). To further address this issue, two-step fine adjustment procedures are carried using Eq.~(\ref{eq_fine_tuning_first}) and (\ref{eq_fine_tuning_second}) based on the both robotic tracking information and B-mode images. The tracking information is used to fully overlap the last frame of the first partition and the first frame of the second partition. Then an image-based in-plane adjustment is performed to overlap the two centroids of the target artery in the overlapped two frames. The 3D result after fine adjustment is shown in Fig.~\ref{Fig_3D_compesated}~(c), where the stitching has been successfully compensated. This will further enable autonomous diagnosis.

\section{Discussion}
Precise repositioning of US probe is one of the crucial techniques for guaranteeing accurate and complete geometry of target anatomies when the object's pose is changed during scanning. This study first proposed a systematic pipeline for RUSS to monitor and further compensate for the potential object motion only on a depth camera. The proposed motion-aware RUSS performs boldly in our setup of experiments (on a human-like arm phantom). However, there are still some limitations that need to be discussed. First, in this work, we have so far considered the rigid motion of object. The articulated motion and deformation of object are existing as well for US scanning. However, the articulated motion could be consider as multiple rigid motion around joints. Regarding the force-induced deformation, one can apply stiffness-based approach to achieve zero-compress volume~\cite{jiang2021deformation}. In addition, the ICP-based compensation method was designed for visible object motions during US scans. The tremor or vessel motion~\cite{chen20163d} have not been considered. However, due to the use of the compliant controller, the small and fast physiological motion can be partially adapted by the controller. Thus, compared with free-hand US acquisition, the effect of such small motions is not aggravated in our setup. At this moment, the presented RUSS can be used for the potential applications on rigid body parts, i.e., forearm, upper arm, thigh, shank or whole chest. By further considering articulated motion~\cite{jiang2022towards} and integrating deformation correction approach~\cite{jiang2021deformation} and \revision{non-rigid registration~\cite{zhang2021reliable} in the future, the system has the potential to provide reliable US images in clinical practice.} \revision{It is also noteworthy that the fine-tuning process theoretically cannot improve the accuracy of the 3D geometry of the vessel of interest. The objective of the fine-tuning process is to improve the visualization performance and further enable the autonomous diagnosis of PAD by estimating the local diameter variations in future work.}

\section{Conclusion}
\par
This work introduces a motion-aware robotic US system only based on an RGB-D camera. This system enables achieving accurate and consistent 3D images of target anatomy when the object is changed during US scans. In this way, the proposed system possesses the advantages of free-hand US (flexibility) and robot US (accuracy and stability) at the same time. To this end, the scanning trajectory on skin is generated by registering the surface point cloud obtained from the camera to the CT template with annotated objects under skin. To monitor object movements, dice coefficients are computed based on the binary masks of object segmented from two different frames using a well-trained UNet-VGG16 model. Once the object is moved, the surface registration-based approach is employed to automatically update the sweep trajectory and resume the scan from the break point to seamlessly accomplish the planned sweep. Since the ICP-based registration may result in sub-optimal contact condition between probe and object after repositioning, a confidence-based probe orientation optimization approach is employed. \revision{Finally, to accurately display 3D objects without significant stitching gaps as~\cite{jiang2021motion}, a step-wise fine adjustment procedure is carried out based on both tracking data and B-mode images [see Fig.~\ref{Fig_3D_compesated}~(c)].} Such improvement we believe can make RUSSs more robust and thus, bring them closer to clinical use. Furthermore, it enables the research topic of automatic examination and diagnosis, which is significantly meaningful in pandemic and undeveloped countries and areas.



\section*{ACKNOWLEDGMENT}
The authors would like to acknowledge the Editor-In-Chief, Associate Editor and anonymous reviewers for their contributions to the improvement of this article.

\bibliographystyle{IEEEtran}
\bibliography{IEEEabrv,references}

\vspace{-0.3cm}
\begin{IEEEbiography}[{\includegraphics[width=1in,height=1.25in,clip,keepaspectratio]{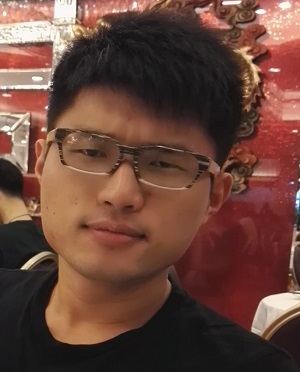}}]
{Zhongliang Jiang} (Graduate Student Member, IEEE) received the M.Eng. degree in Mechanical Engineering from the Harbin Institute of Technology, Shenzhen, China, in 2017. From January 2017 to July 2018, he worked as a research assistant at the Shenzhen Institutes of Advanced Technology (SIAT) of the Chinese Academy of Science (CAS), Shenzhen, China. He is currently working toward the Ph.D. degree in Computer Science at the Technical University of Munich, Munich, Germany.

His research interests include medical robotics, robotic learning, human-robot interaction, and robotic ultrasound, computer vision.
\\
\end{IEEEbiography}

\vspace{-1cm}
\begin{IEEEbiography}
[{\includegraphics[width=1in,height=1.25in,clip,keepaspectratio]{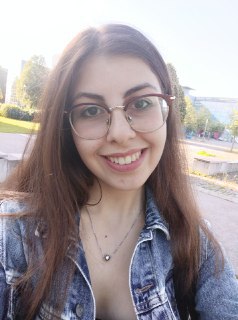}}]
{Nehil Danis} received the B.Sc. degree in Computer Science from the Hacettepe University, Ankara, Turkey, in 2018 and the M.Sc. degree in Computer Science from the Technical University of Munich, Munich, Germany, in 2021. She is currently working as a software engineer at Brainlab AG, Munich, Germany.

Her research interests include 3D geometry processing, visual SLAM, and robotic manipulation.
\\ 
\end{IEEEbiography}

\vspace{-1cm}
\begin{IEEEbiography}[{\includegraphics[width=1in,height=1.25in,clip,keepaspectratio]{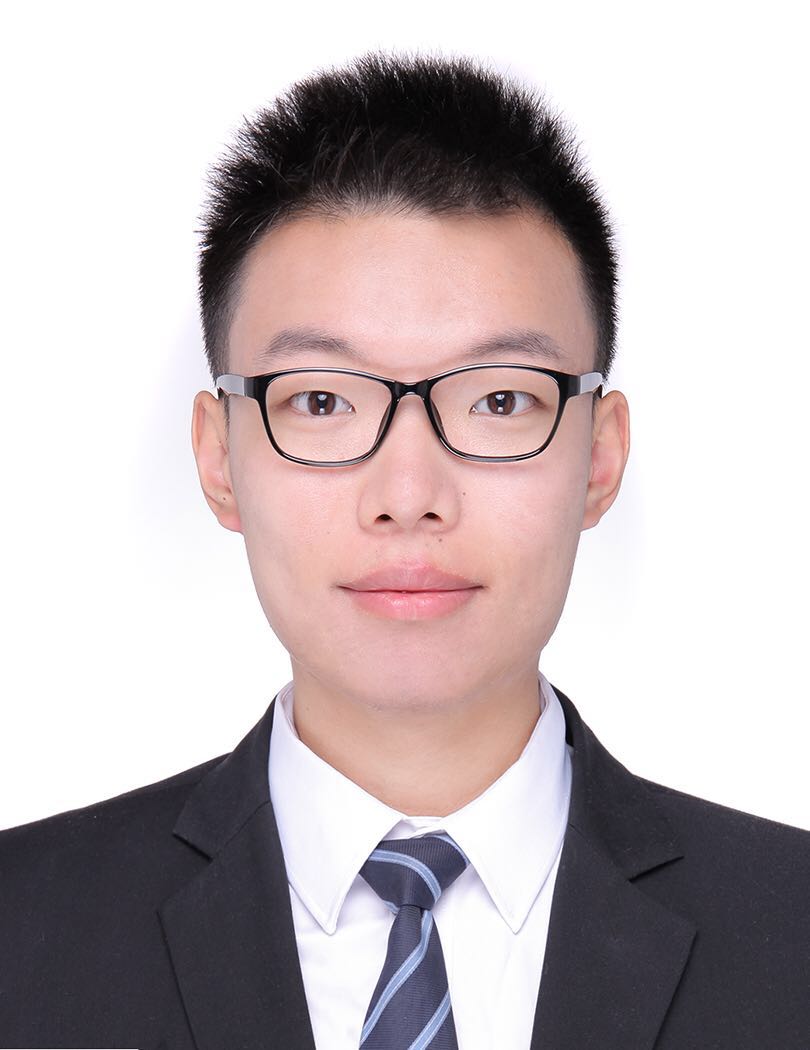}}]
{Yuan Bi} received the B.Sc. degree in Mechatronics from the Tongji University, Shanghai, China, 2017, and M.Sc. degree in Robotics, Cognition, Intelligence from the Technical University of Munich, Munich, Germany, 2021. He is currently working toward the Ph.D. degree in Computer Science with the Technical University of Munich, Munich, Germany.

His research interests include robot manipulation, robotic learning, and computer vision.
\\ 
\end{IEEEbiography}

\vspace{-1cm}
\begin{IEEEbiography}
[{\includegraphics[width=1in,height=1.25in,clip,keepaspectratio]{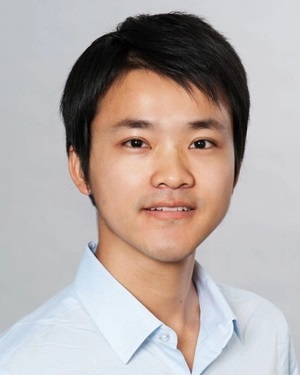}}]
{Mingchuan Zhou} received Ph.D. degree in computer science from the Technical University of Munich, Munich, Germany, in 2020. He was the visiting scholar at Laboratory for Computational Sensing and Robotics, Johns Hopkins University, USA, 2019. He was a joint postdoc at Institute of Biological and Medical Imaging (IBMI) of the Helmholtz Center Munich and Chair for Computer Aided Medical Procedures Augmented Reality (CAMP) at the Technical University of Munich from 2019 to 2021. He is currently an Assistant Professor leading multi scale robotic manipulation lab for agriculture in Zhejiang University. His research interests include the autonomous system, agricultural robotics, medical robotics, and image processing.
\\ 
\end{IEEEbiography}

\vspace{-1cm}
\begin{IEEEbiography}
[{\includegraphics[width=1in,height=1.25in,clip,keepaspectratio]{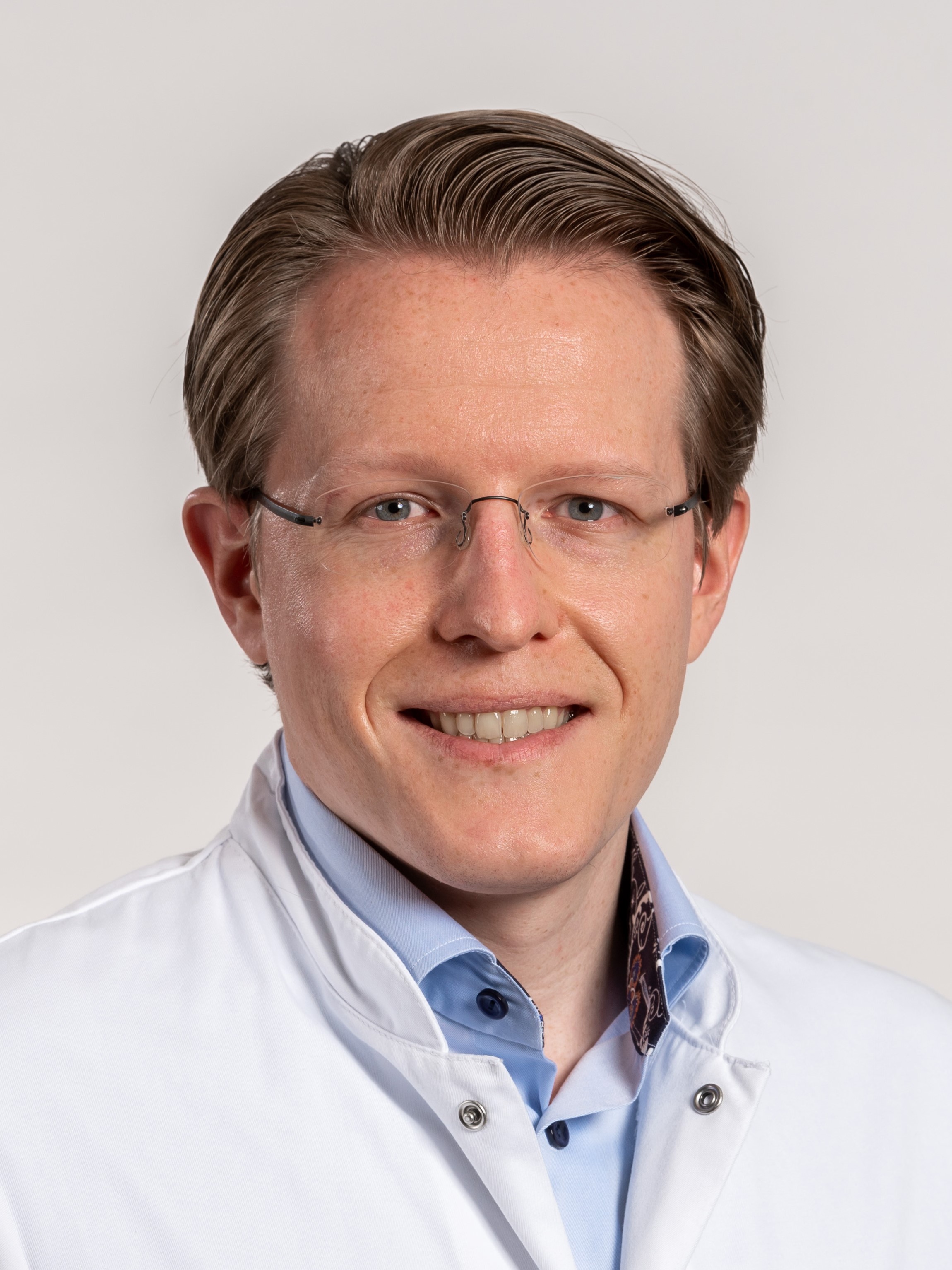}}]
{Markus Kroenke} studied physics and medicine at the Ludwig Maximilians University, Munich and the Technical University of Munich, Germany. He is board certified in Nuclear Medicine and Attending Physician at the Department of Radiology and Nuklearmedicine at the German Heart Center Munich, Technical University of Munich.

His main research focus is multimodal imaging.
\\ 
\end{IEEEbiography}

\vspace{0.3cm}
\begin{IEEEbiography}[{\includegraphics[width=1in,height=1.25in,clip,keepaspectratio]{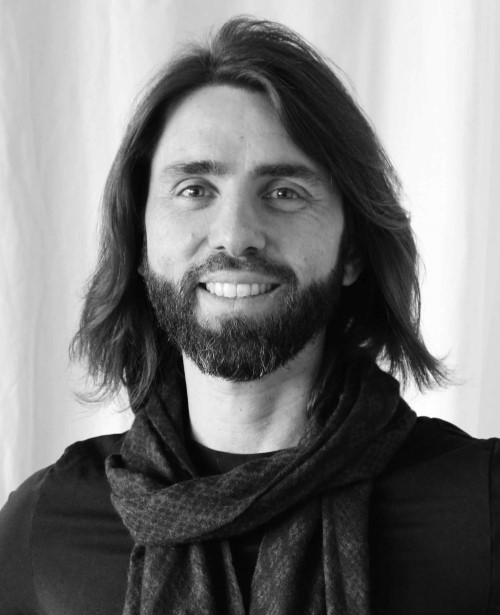}}]
{Thomas Wendler} received an Electrical Engineering degree at Universidad Técnica Federico Santa María, Chile. Subsequently, he obtained a M.Sc. degree in Biomedical Engineering and a degree in Computer Science from the Technical University of Munich, Germany. He is currently the Vice-Director of the chair for Computer-Aided Medical Procedures and Augmented Reality (CAMP) at the Technical University of Munich (TUM), CEO of ScintHealth GmbH, and interim CTO of SurgicEye GmbH.

His main research focus is the translation of novel computer-aided intervention tools and data-driven medical image analysis methods for clinical decision support into clinical applications. 

\\ 
\end{IEEEbiography}

\vspace{-1cm}
\begin{IEEEbiography}[{\includegraphics[width=1in,height=1.25in,clip,keepaspectratio]{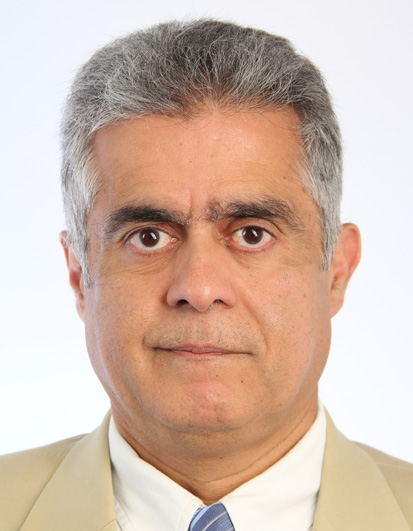}}]
{Nassir Navab} (Fellow, IEEE) received the Ph.D. degree in computer and automation with INRIA, and the University of Paris XI, Paris, France, in 1993.

He is currently a Full Professor and the Director of the Laboratory for Computer-Aided Medical Procedures with the Technical University of Munich, Munich, Germany, and an adjunct professor at Johns Hopkins University, Baltimore, MD, USA. He has also secondary faculty appointments with the both affiliated Medical Schools. He enjoyed two years of a Postdoctoral Fellowship with the MIT Media Laboratory, Cambridge, MA, USA, before joining Siemens Corporate Research (SCR), Princeton, NJ, USA, in 1994. 

Dr. Navab is a fellow of the Academy of Europe, MICCAI, IEEE, and Asia-Pacific Artificial Intelligence Association (AAIA). He was a Distinguished Member and was the recipient of the Siemens Inventor of the Year Award in 2001, at SCR, the SMIT Society Technology award in 2010 for the introduction of Camera Augmented Mobile C-arm and Freehand SPECT technologies, and the ``$10$ years lasting impact award" of IEEE ISMAR in 2015. He is the author of hundreds of peer-reviewed scientific papers, with more than 54,400 citations and enjoy an h-index of 104 as of August 11, 2022. He is the author of more than thirty awarded papers including 11 at MICCAI, 5 at IPCAI, and three at IEEE ISMAR. He is the inventor of 50 granted US patents and more than 50 International ones.
\\ 
\end{IEEEbiography}

\end{document}